\documentclass[10pt,twocolumn,letterpaper]{article}

\usepackage{cvpr}      
\usepackage{multirow} 
\usepackage{tabularx} 
\usepackage{booktabs}
\definecolor{cvprblue}{rgb}{0.21,0.49,0.74}
\usepackage[pagebackref,breaklinks,colorlinks,allcolors=cvprblue]{hyperref}

\def\paperID{21} 
\def\confName{CVPR}
\def\confYear{2026}

\title{LEVERAGING IMAGE EDITING FOUNDATION MODELS FOR DATA-EFFICIENT CT
METAL ARTIFACT REDUCTION}

\author{
Ahmet Rasim Emirdagi \qquad Süleyman Aslan \qquad Mısra Yavuz \qquad Görkay Aydemir\\
Yunus Bilge Kurt \qquad Nasrin Rahimi \qquad Burak Can Biner \qquad M.\ Akın Yılmaz\\[0.5em]
Codeway AI Research
}

\begin{document}
\maketitle
\begin{abstract}
Metal artifacts from high-attenuation implants severely degrade CT image quality, obscuring critical anatomical structures and posing a challenge for standard deep learning methods that require extensive paired training data. We propose a paradigm shift: reframing artifact reduction as an in-context reasoning task by adapting a general-purpose vision-language diffusion foundation model via parameter-efficient Low-Rank Adaptation (LoRA). By leveraging rich visual priors, our approach achieves effective artifact suppression with only 16 to 128 paired training examples reducing data requirements by two orders of magnitude. Crucially, we demonstrate that domain adaptation is essential for hallucination mitigation; without it, foundation models interpret streak artifacts as erroneous natural objects (e.g., waffles or petri dishes). To ground the restoration, we propose a multi-reference conditioning strategy where clean anatomical exemplars from unrelated subjects are provided alongside the corrupted input, enabling the model to exploit category-specific context to infer uncorrupted anatomy. Extensive evaluation on the AAPM CT-MAR benchmark demonstrates that our method achieves state-of-the-art performance on perceptual and radiological-feature metrics . This work establishes that foundation models, when appropriately adapted, offer a scalable alternative for interpretable, data-efficient medical image reconstruction. Code is available at \url{https://github.com/ahmetemirdagi/CT-EditMAR}.
\end{abstract}    
\section{Introduction}
\label{sec:intro}

Computed Tomography (CT) is a cornerstone of modern diagnostic imaging due to its speed, accessibility, and high spatial resolution. However, high-density metallic implants such as orthopedic hardware and dental fillings introduce severe image degradations known as metal artifacts. These artifacts arise from beam hardening, photon starvation, and scattering, manifesting as streaking and shadowing patterns~\cite{bolstad2022ct, zhang2023advances}. Consequently, soft-tissue regions adjacent to metal implants are often rendered non-diagnostic, limiting clinical utility in routine and post-operative scenarios~\cite{chang2023virtual}.

Early metal artifact reduction (MAR) approaches focused on acquisition-level strategies, including gantry angle adjustment, increased tube voltage, and dual-energy CT~\cite{zhang2023advances}. While effective in specific scenarios, these methods are limited by clinical constraints regarding radiation dose, noise, and scan protocols. Subsequent projection-domain correction methods explicitly target metal-corrupted sinogram regions using interpolation or inpainting strategies~\cite{kalender1987reduction,meyer2010nmar}. Iterative reconstruction frameworks further incorporate physical imaging models of polychromatic X-ray spectra and noise statistics to improve consistency~\cite{deman2001iterative,wang1996iterative}. However, these methods rely heavily on accurate metal segmentation and access to raw projection data. Moreover, model mismatch and severe photon starvation often lead to residual artifacts and secondary distortions, limiting robustness with large implants~\cite{gjesteby2016metal}.

Recent research has shifted toward data-driven deep learning approaches, which learn artifact suppression directly from large paired datasets. The most capable methods operate in the dual domain, enforcing consistency between raw sinogram measurements and reconstructed image space~\cite{lin2019dudonet,indudonet}. Although effective, such approaches carry a critical clinical limitation: raw sinogram data is rarely preserved or accessible in hospital PACS systems, where only reconstructed CT volumes are stored. This reduces the number of dual-domain methods for research of curiosities with limited translational value.

Image-domain methods avoid this constraint, but have historically underperformed their dual-domain counterparts as they lack the geometric consistency enforced by sinogram supervision. Both paradigms, however, share a more fundamental bottleneck: their reliance on tens of thousands of paired training examples, which demands significant computational infrastructure and must be repeated for each scanner protocol and implant type.

To address this data bottleneck, recent works have explored unsupervised and GAN-based architectures~\cite{liao2019adn, wang2018cgan} and generative approaches such as diffusion models~\cite{song2022solving}. While these methods reduce annotation costs, they often suffer from anatomical hallucinations, training instability, and prohibitive inference times. There remains a critical need for a framework combining data efficiency with structural fidelity.

Foundation models offer a natural path toward this dual requirement. Large-scale foundation models pretrained on internet-scale visual corpora have demonstrated remarkable generalization across image understanding and generation tasks~\cite{radford2021clip,rombach2022ldm,openai2023gpt4v}. Their ability to encode rich structural priors like lighting, texture, geometry, and semantic contexts from billions of natural images has motivated a wave of medical imaging adaptations. Models such as MedSAM~\cite{ma2024medsam} and BiomedCLIP~\cite{zhang2023biomedclip} show that foundation model representations transfer surprisingly well to clinical domains even without medical pretraining, while SAM-Med2D~\cite{cheng2023samm} and related work demonstrate competitive segmentation with minimal labelled data.

Critically, this transfer does not require full retraining: parameter-efficient fine-tuning (PEFT) methods~\cite{dft_peft, foura_peft, waveft_peft, dora_peft, vera_peft}, Low-Rank Adaptation (LoRA)~\cite{loraref} in particular, adapt billion-parameter models to new tasks by injecting small low-rank weight residuals while keeping the backbone frozen. LoRA has been shown to match full fine-tuning quality on downstream tasks with as few as 1--5\% of the trainable parameters, and its compact checkpoint size makes it practical to train and deploy separate adapters per scanner protocol or implant type. Recent work has shown that pre-trained image editing foundation models can be adapted for few-shot image restoration via parameter-efficient fine-tuning within different domains by requiring just only 16 to 128 paired examples for tasks such as denoising, deraining, and dehazing or video interpolation ~\cite{yilmaz2026edit2restore, rahimi2026edit2interp}. Building on this emerging paradigm, we investigate whether such adaptation can extend to the clinically important and structurally challenging problem of CT metal artifact reduction. Despite these advances, the application of generative foundation models to CT artifact correction remains largely unexplored: prior MAR methods are purpose-built networks trained from scratch, inheriting none of the structural priors embedded in large pretrained models. We address this gap by adapting Qwen-Image-Edit 2511~\cite{qwenref}, a 20\,B-parameter vision-language diffusion model trained for general image editing, to CT metal artifact removal via LoRA fine-tuning on as few as 16 paired slices.

Our experiments on the AAPM CT-MAR benchmark~\cite{haneda2025aapm} show that the proposed approach can suppress metal artifacts effectively with as few as 16 paired training examples, which is roughly two orders of magnitude less than what conventional supervised methods typically require, while preserving anatomical plausibility and achieving competitive perceptual quality scores against specialized baselines.

Our contributions are:
\begin{itemize}
\item We are the first to adapt a general-purpose image editing foundation model for CT metal artifact reduction, demonstrating that rich visual priors transfer effectively to CT without domain-specific pretraining.
\item We propose a reference-conditioned fine-tuning strategy that incorporates clean patient-matched CT slices as auxiliary inputs, enabling anatomy-aware restoration without sinogram access.
\item We achieve state-of-the-art perceptual quality on the AAPM CT-MAR benchmark using only 16–128 training pairs, establishing a new data-efficiency frontier for clinical MAR deployment.
\end{itemize}
\begin{figure}[t]
    \centering
    \includegraphics[width=\columnwidth]{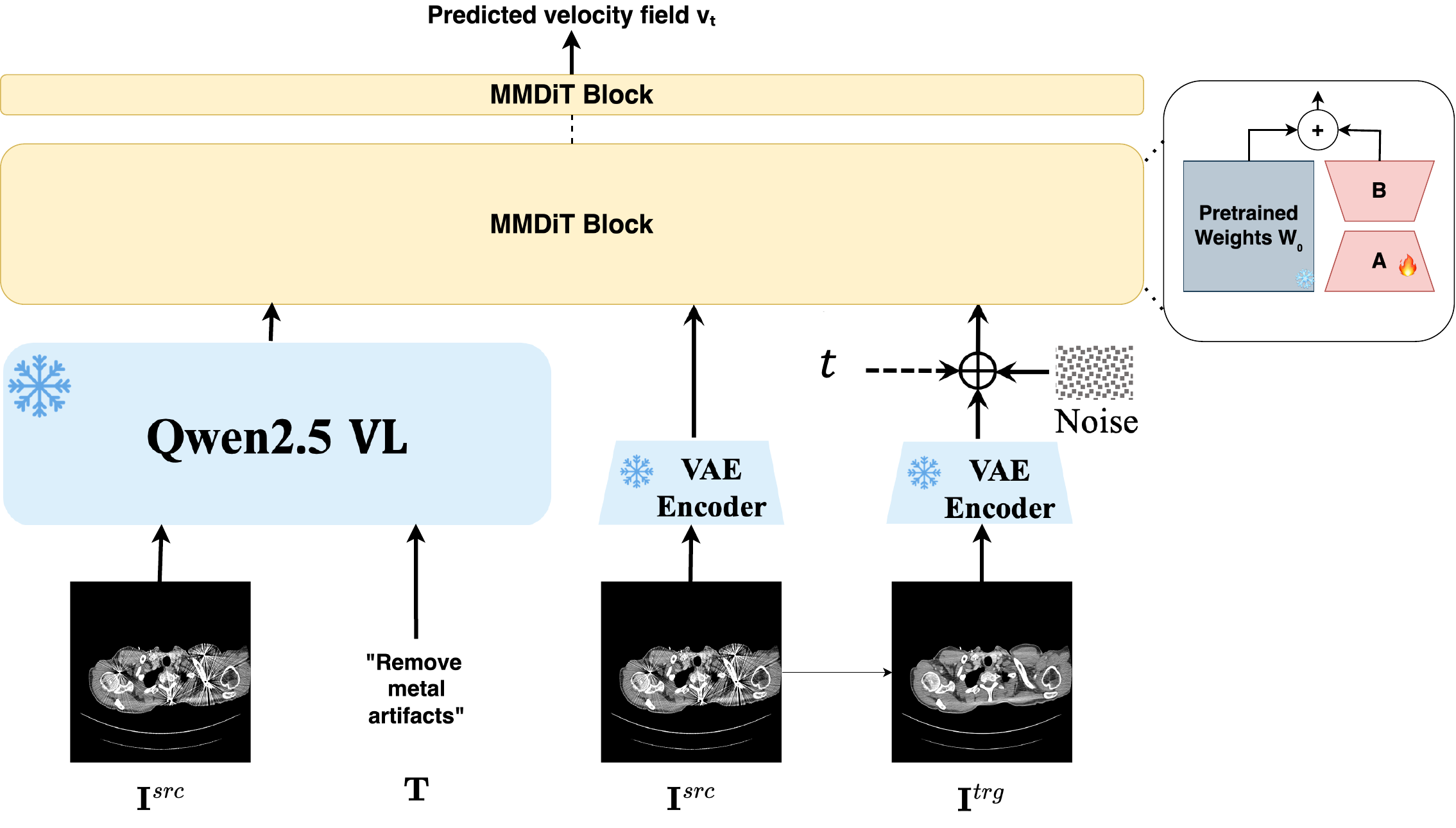}
    \captionsetup{font=small}
    \caption{Overview of our framework}
    \label{fig:pipeline}
    \vspace{-6pt}
\end{figure}

\section{Methods}
\label{methods}

\subsection{Problem Formulation}
\label{ssec:problem_def}

To leverage the generative priors of instruction-conditioned image editing
foundation models, we formulate metal artifact reduction as a
\textit{conditional image restoration} task. As illustrated in
Figure~\ref{fig:pipeline}, given a metal-corrupted CT slice
$\mathbf{I}^{src} \in \mathbb{R}^{H \times W}$, a set of $K$ clean
reference slices $\mathbf{R} = \{r_1, \ldots, r_K\}$ drawn from clean CT slices from other patients in the training set (unrelated subjects), used as style/anatomy exemplars, and a textual instruction
$\mathbf{T}$ (e.g., \textit{``Remove metal artifacts''}), the goal is to
estimate an artifact-reduced slice $\hat{\mathbf{I}}^{trg}$ via a
parameterized mapping
\begin{equation}
    \hat{\mathbf{I}}^{trg} = \mathcal{F}_{\theta}
        \!\left(\mathbf{I}^{src},\, \mathbf{R},\, \mathbf{T}\right),
\end{equation}
where $\mathcal{F}_{\theta}$ should suppress metal-induced streaking and
shading while preserving anatomical structures in unaffected regions and
reconstructing artifact-corrupted areas consistent with the surrounding
context.  The reference set $\mathbf{R}$ provides patient-specific
anatomical context without requiring access to raw sinogram data, making
the formulation directly applicable to standard clinical PACS images.
When no reference slices are available, the framework reduces to the
reference-free special case $\mathcal{F}_{\theta}(\mathbf{I}^{src}, \mathbf{T})$.

\subsection{Model and Parameter-Efficient Adaptation}
\label{ssec:lora}

To approximate the restoration function $\mathcal{F}_\theta$, we adopt
Qwen-Image-Edit~\cite{qwenref}, an instruction-conditioned image editing
model built on a 20-billion parameter diffusion transformer (DiT)
pretrained at scale on natural images. Unlike formulations that treat
restoration purely as pixel-wise regression, Qwen-Image-Edit frames
restoration as a conditional generation problem: given a corrupted input
and a textual instruction describing the desired correction, the model
generates a restored image by conditioning its generative backbone on
both the visual input and the instruction. This allows the model to
exploit learned structural priors to fill artifact-corrupted regions in a
manner that is globally consistent with the surrounding anatomy.

Fine-tuning all parameters of this backbone is computationally expensive
and prone to overfitting in the low-data regimes typical of paired
clinical CT datasets. We therefore employ Low-Rank Adaptation
(LoRA)~\cite{loraref}, a parameter-efficient fine-tuning strategy that
injects trainable low-rank update matrices into the frozen pretrained
backbone.

Formally, for a pretrained weight matrix
$\mathbf{W}_0 \in \mathbb{R}^{d \times k}$ (e.g., in an attention
projection layer), LoRA freezes $\mathbf{W}_0$ and parameterizes the
weight update as $\Delta\mathbf{W} = \mathbf{B}\mathbf{A}$, where
$\mathbf{B} \in \mathbb{R}^{d \times r}$,
$\mathbf{A} \in \mathbb{R}^{r \times k}$, and $r \ll \min(d, k)$ is
the adaptation rank.  The forward pass becomes
\begin{equation}
    h = \mathbf{W}_0 x + \frac{\alpha}{r}\,\mathbf{B}\mathbf{A}x,
\end{equation}
where $x$ is the layer input and $\alpha$ is a constant scaling factor.
Following standard practice, $\mathbf{A}$ is initialized from a random
Gaussian distribution and $\mathbf{B}$ is initialized to zero, so that
$\Delta\mathbf{W} = 0$ at the start of training and the pretrained
representations are preserved; only $\mathbf{A}$ and $\mathbf{B}$ are
optimized.  This reduces the number of trainable parameters by orders of
magnitude relative to full fine-tuning, enabling stable adaptation from
as few as 16 paired training examples while retaining the rich visual
priors of the pretrained backbone.

\subsection{Reference Conditioning Strategies}
\label{sec:ref_conditioning}

To supply the model with clean anatomical context, we assemble a
reference set $\mathcal{R} = \{r_1, \ldots, r_5\}$ of five metal-free
CT slices drawn from different patients in a held-out reference pool.
We investigate two strategies for incorporating $\mathcal{R}$ into the
single-step inference call.

\paragraph{Multi-image conditioning}
The Qwen-Image-Edit backbone is a vision-language model that natively
accepts a sequence of images as a multi-image prompt.
We exploit this capability by passing the artifact-corrupted slice and
all five references as a single six-image list:
\begin{equation}
    \hat{I} = \mathcal{D}\!\left(
        F_\theta\!\bigl([\mathcal{E}(I^{\mathrm{src}}),\,
                          \mathcal{E}(r_1), \ldots, \mathcal{E}(r_5)],\;
                         T_{\mathrm{multi}}\bigr)
    \right),
\end{equation}
where the prompt $T_{\mathrm{multi}}$ explicitly labels each image
(``Image~1 is the CT scan with metal artifacts \ldots
Images~2--6 are clean reference CTs'').
All six images are processed in a single forward pass through the
transformer; the cross-image attention inherent to the VLM architecture
allows the model to draw structural cues directly from the references.

\paragraph{Grid-concatenation conditioning}
As a simpler baseline, we tile the six images into a $2 \times 3$
pixel grid (each cell resized to $256 \times 256$) and pass the
resulting $512 \times 768$ composite as a single RGB image.
The prompt $T_{\mathrm{grid}}$ describes the layout (``The input is a
2×3 grid. Top-left is the CT scan \ldots the other five cells are clean
reference CTs''), relying on the model's spatial understanding rather
than explicit multi-image tokens.
No architectural change is required; the grid is treated as an ordinary
single-image edit.

\subsection{Self-ensembling}
\label{sec:ensembling}
Diffusion models are stochastic at inference time.
To reduce this variance for our best configuration, we run the
multi-image model $M\!=\!10$ times with different random seeds and
average the resulting pixel-space outputs:
\begin{equation}
    \hat{I}_{\mathrm{ens}} = \frac{1}{M}\sum_{n=1}^{M}
        \mathcal{D}\!\left(
            F_\theta\!\bigl([\mathcal{E}(I^{\mathrm{src}}),\,
                              \mathcal{E}(r_1), \ldots, \mathcal{E}(r_5)],\;
                             T_{\mathrm{multi}},\; s_n\bigr)
        \right),
\end{equation}
where $s_n$ denotes the seed for run $n$.
This is a test-time technique independent of the conditioning strategy.

\begin{table*}[t]
\centering
\caption{
    Quantitative comparison on the 1{,}000-sample CT-MAR test set.
    \textbf{Bold}: best among trained models.
    \underline{Underline}: second best among trained models.
    VAE ceiling is not a trained MAR model — it encodes and decodes clean GT
    images without any diffusion, representing the theoretical upper bound
    imposed by the VAE bottleneck.
}
\label{tab:main_results}
\setlength{\tabcolsep}{3pt}
\begin{tabular}{llcccccc}
\toprule
& Method & PSNR (dB)$\uparrow$ & SSIM$\uparrow$ & FID$\downarrow$ & FID-Rad$\downarrow$ & LPIPS$\downarrow$ & LPIPS-Rad$\downarrow$ \\
\midrule
\multirow{3}{*}{Baselines}
& ADN~\cite{liao2019adn}              & 31.38 & 0.8507 & 59.09 & 0.0600 & 0.1841 & 0.1708 \\
& RISE-MAR~\cite{risemar}             & 32.61 & 0.8514 & 66.12 & 0.0673 & 0.1776 & 0.1744 \\
& OSCNet+~\cite{oscnet+}        & \textbf{37.74} & \underline{0.8832} & 31.60 & 0.0241 & 0.1531 & 0.0931 \\
& Qwen-Image-Edit Zero-Shot       & 4.21 & 0.1325 & 193.66 & 0.832 & 0.3706 & 0.7609 \\
\midrule
\multirow{3}{*}{Ours}
& LoRA ($N$=128, $r$ = 64, no refs)             & 31.61 & 0.8717 & 5.81 & 0.0038 & 0.1062 & 0.0437 \\
& LoRA ($N$=128, $r$ = 64, multi-refs)           & 34.60 & 0.8829 & \textbf{4.91} & \textbf{0.0031} & \textbf{0.0926} & \textbf{0.0371} \\
& LoRA ($N$=128, $r$ = 64, multi-refs, ens-10)   & \underline{35.51} & \textbf{0.8966} & \underline{5.33} & \underline{0.0037} & \underline{0.0972} & \underline{0.0391} \\
\midrule
\multirow{1}{*}{Upper bound}
& VAE ceiling          & 38.57 & 0.9313 & 0.74 & 0.0002 & 0.0375 & 0.0080 \\
\bottomrule
\end{tabular}
\end{table*}

\subsection{Training}
\label{sec:training}

\paragraph{Data preparation.}
Training pairs are derived from our CT-MAR dataset of paired artifact-corrupted
and clean CT volumes.
Each slice is windowed to the HU range $[-500, 1300]$ and linearly mapped to
an 8-bit PNG ($512 \times 512$ pixels). The upper bound of 1300HU intentionally saturates metal implant
voxels, which typically exceed 3000HU, to a uniform maximum
intensity; since the restoration target is the artifact-corrupted
\emph{soft tissue} rather than the implant itself, no clinically
relevant information is lost by this saturation.

The noisy slice serves as the edit image and the clean slice as the
reconstruction target.
For the reference-conditioned variants (Section~\ref{sec:ref_conditioning}),
five additional metal-free reference slices from other patients are
appended to each training example.
We study four training-set sizes: $N \in \{16, 32, 64, 128\}$ pairs.

\paragraph{Loss function}
We adopt the flow-matching supervised fine-tuning objective
(FlowMatchSFT)~\cite{lipman2022flow,esser2024scaling}.
For a training pair $(x_0, x_1)$ where $x_1$ is the clean latent and
$x_0$ is pure noise, the training signal at interpolation time $t$ is:

\begin{equation}
    \begin{split}
        \mathcal{L} = \mathbb{E}_{t,\, x_0,\, x_1}
        \left\|
            v_\theta\!\left(x_t,\, t,\, \mathcal{E}(I^{\mathrm{src}}),\, T\right)
            - (x_1 - x_0)
        \right\|_2^2, \\
        x_t = (1-t)\,x_0 + t\,x_1,
    \end{split}
\end{equation}

where $v_\theta$ is the transformer's predicted velocity field
and $\mathcal{E}(I^{\mathrm{src}})$ is the VAE-encoded noisy input
provided as edit conditioning.
All backbone parameters are frozen; only the LoRA weights are updated.

\paragraph{Optimisation}
All models are trained with AdamW (learning rate $10^{-4}$, no scheduler)
for a fixed budget of 750 gradient steps.
To equalize the step budget across dataset sizes, the dataset is repeated
$\lceil 750 / N \rceil$ times so that each model performs the same total
number of parameter updates regardless of $N$.
Gradient checkpointing is enabled throughout.
Training is conducted on two NVIDIA H100 GPUs (one model per GPU, sequential
jobs per GPU).

\paragraph{LoRA configuration}
LoRA adapters are applied to the DiT backbone at all attention query, key,
and value projections in both the image stream
($\mathtt{to\_q}$, $\mathtt{to\_k}$, $\mathtt{to\_v}$,
 $\mathtt{to\_out.0}$)
and the cross-stream modulation layers
($\mathtt{add\_q\_proj}$, $\mathtt{add\_k\_proj}$, $\mathtt{add\_v\_proj}$,
 $\mathtt{to\_add\_out}$),
as well as the image and text MLP/modulation heads
($\mathtt{img\_mlp.net.2}$, $\mathtt{img\_mod.1}$,
 $\mathtt{txt\_mlp.net.2}$, $\mathtt{txt\_mod.1}$) meaning 12 module types in total.
The rank $r$ is ablated over $\{16, 32, 64, 128\}$;
all other LoRA hyperparameters use the DiffSynth-Studio defaults.

\begin{figure*}[t]
    \centering
    \includegraphics[width=0.9\textwidth]{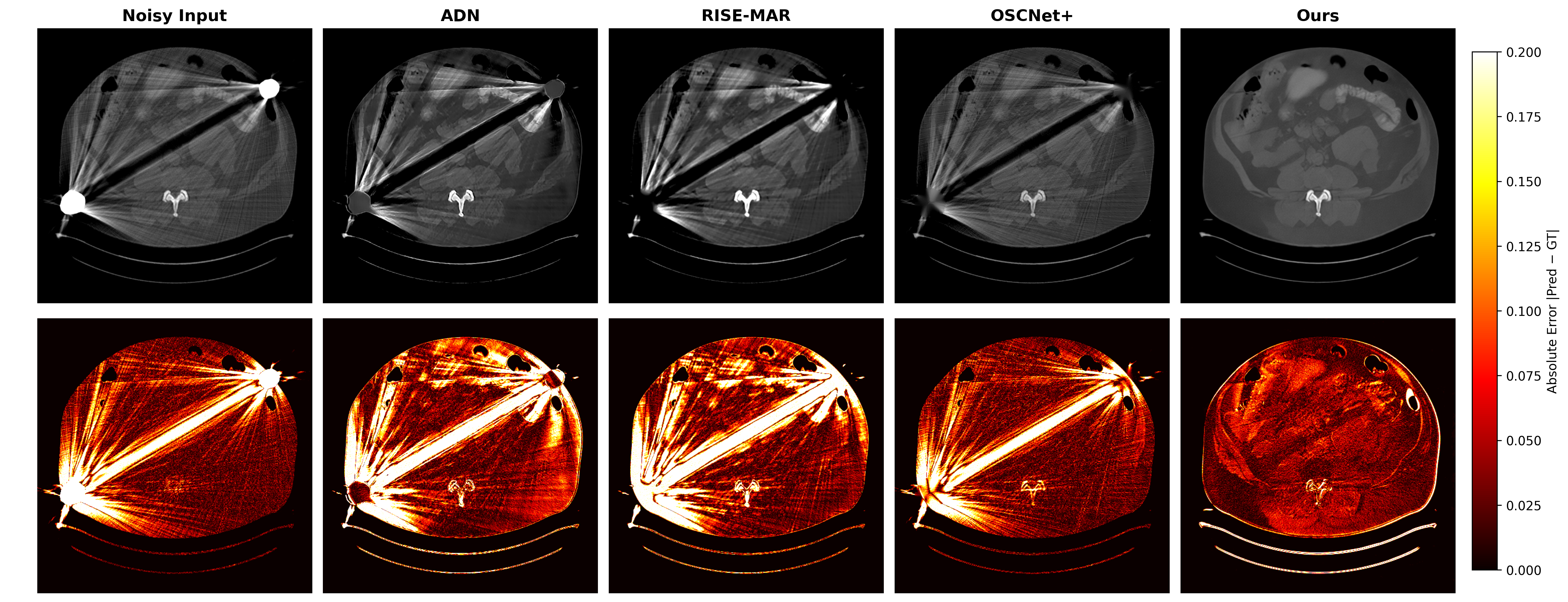}
    \caption{Visual comparison of artifact reduction methods. 
    (a) Noisy inputs, (b) ADN, (c) Rise-MAR, (d) OSCNet+ and (e) Our reconstructions. The bottom row shows the absolute error maps relative to the ground truth. 
    Note how our method significantly reduces streak artifacts while preserving anatomical structure.}
    \label{fig:method_comparison}
\end{figure*}

\section{Experiments}

\subsection{Dataset}
\label{sec:dataset}

This work utilizes the AAPM CT Metal Artifact Reduction (CT-MAR) Grand Challenge benchmark \cite{haneda2025aapm, aapmBenchmarkTool}. The dataset is generated using a hybrid simulation framework \cite{peters2025hybrid} combining publicly available clinical CT images \cite{yan2018deeplesion, goren2017uclh}, clinical images from Massachusetts General Hospital, and simulated metallic implants using the CatSim simulator in the XCIST toolkit \cite{wu2022xcist}. Notably, the clinical CT images underlying the AAPM benchmark are drawn primarily from the NIH DeepLesion dataset \cite{yan2018deeplesion}, ensuring that the patient anatomical distribution overlaps with the training domain of several baselines evaluated in Section ~\ref{sec:baselines}.

For evaluation we hold out 1\,000 slices drawn from the ten body categories, with no overlap between training and test
indices.
The training subsets consist of $N \in \{16, 32, 64, 128\}$ pairs
sampled uniformly from the remaining slices.

All images are windowed to the HU range $[-500, 1300]$ and converted to
8-bit PNG for training and evaluation. For experiments with reference conditioning, up to five metal-free target slices from the \emph{same anatomical category} (but different people) are associated with each test sample.

\subsection{Baselines}
\label{sec:baselines}

We compare against three representative image-domain MAR methods, all
evaluated using their publicly released pretrained weights:

\noindent\textbf{ADN~\cite{liao2019adn}}
Artifact Disentanglement Network, trained on the DeepLesion dataset.
Inputs are converted from HU to attenuation coefficients
$(\text{att} = \text{HU}/1000 \times 0.192 + 0.192)$,
clipped to $[0, 0.5]$, and normalized to $[-1, 1]$ before inference.

\noindent\textbf{RISE-MAR~\cite{risemar}}
A residual iterative self-enhancement network pretrained on body CT.
HU values are normalized to $[0,1]$ using the model's training range
$[-1024, 3072]$.

\noindent\textbf{OSCNet / OSCNet+~\cite{oscnet, oscnet+}}
Orientation-shared convolutional network with a sinogram-consistency prior
approximated by image-domain inpainting (OpenCV TELEA~\cite{telea2004})
of the metal region.
Trained on the SynDeepLesion benchmark.
The metal mask is derived by thresholding the raw HU volume at
$\text{HU} > 2000$, dilated with a $7 \times 7$ elliptical kernel.

Since the AAPM CT-MAR benchmark itself derives its patient images from DeepLesion \cite{yan2018deeplesion}, ADN and OSCNet+ \cite{liao2019adn, oscnet, oscnet+} operate on a familiar anatomical distribution; the primary difference lies in the metal artifact simulation pipeline rather than the source patient anatomy, making the comparison fair.

\noindent\textbf{No-LoRA (zero-shot).}
To quantify the contribution of domain adaptation, we additionally evaluate
the base Qwen-Image-Edit model with no fine-tuning whatsoever, applying the
same input representation and inference pipeline as our
adapted models. This zero-shot configuration produces severely degraded
outputs, confirming that the foundation model's natural-image
priors do not transfer to CT metal artifact reduction without adaptation.

\noindent\textbf{Reference Point}
We additionally report a reference point which is the \textbf{VAE round-trip} (clean GT encoded and decoded without any diffusion or LoRA) as the theoretical upper bound imposed by the VAE bottleneck.

\subsection{Evaluation Metrics}
\label{sec:metrics}

We evaluate on six complementary metrics computed on the full 1{,}000-image test set.
\textbf{Peak Signal-to-Noise Ratio (PSNR)}~\cite{psnr} and \textbf{Structural Similarity Index (SSIM)}~\cite{wang2004ssim} measure pixel-level fidelity on grayscale CT images.
Because pixel agreement alone can be high even when diagnostically relevant structures are over-smoothed or distorted, we additionally report feature-based realism metrics.
\textbf{Fr\'echet Inception Distance (FID)}~\cite{heusel2017fid} assesses distributional similarity using features from an ImageNet-pretrained InceptionV3, and \textbf{Learned Perceptual Image Patch Similarity (LPIPS)}~\cite{zhang2018lpips} measures perceptual distance using an ImageNet-pretrained VGG backbone.
To make these feature distances domain-aware for radiology, we compute \textbf{FID-Rad} and \textbf{LPIPS-Rad} by replacing the feature extractors with networks pretrained on RadImageNet~\cite{mei2022radimagenet}, so the comparisons are performed in a representation space tuned to radiological semantics (e.g., organ boundaries, soft-tissue contrast, and bone structure) rather than natural-image cues.
\begin{table}[t]
\centering
\caption{
    Training set size ablation with rank $r{=}64$ (no reference conditioning). As $r{=}64$ achieved the highest PSNR score on its own ablation.
    \textbf{Bold}: best per metric.
}
\label{tab:size_ablation}
\setlength{\tabcolsep}{4pt}
\renewcommand{\arraystretch}{1.05}
\begin{tabular}{ccccccc}
\toprule
$N$ & PSNR$\uparrow$ & SSIM$\uparrow$ & FID$\downarrow$ & FID-R$\downarrow$ & LPIPS$\downarrow$ & LPIPS-R$\downarrow$ \\
\midrule
16  & 28.25 & 0.8272 & 7.18          & 0.0050          & 0.1232          & 0.0649 \\
32  & 30.14 & 0.8535 & 5.91          & 0.0038          & 0.1033          & 0.0432 \\
64  & 31.41 & 0.8681 & \textbf{5.22} & \textbf{0.0033} & \textbf{0.0951} & \textbf{0.0388} \\
\textbf{128} & \textbf{31.61} & \textbf{0.8717} & 5.81 & 0.0038 & 0.1062          & 0.0437 \\
\bottomrule
\end{tabular}
\end{table}

\begin{table}[t]
\centering
\caption{
    LoRA rank ablation with $N{=}128$ training pairs (no reference conditioning). As $N{=}128$ achieved the highest PSNR score on its own ablation.
    \textbf{Bold}: best per metric.
}
\label{tab:rank_ablation}
\setlength{\tabcolsep}{4pt}
\renewcommand{\arraystretch}{1.05}
\begin{tabular}{ccccccc}
\toprule
$r$ & PSNR$\uparrow$ & SSIM$\uparrow$ & FID$\downarrow$ & FID-R$\downarrow$ & LPIPS$\downarrow$ & LPIPS-R$\downarrow$ \\
\midrule
16  & 25.60 & 0.7603 & 9.45          & 0.0073          & 0.2049          & 0.1320 \\
32  & 26.85 & 0.7792 & 7.85          & 0.0057          & 0.1773          & 0.0985 \\
\textbf{64}  & \textbf{31.61} & \textbf{0.8717} & \textbf{5.81} & \textbf{0.0038} & \textbf{0.1062}          & \textbf{0.0437} \\
128 & 30.49 & 0.8588 & 7.48 & 0.0056 & 0.1076 & 0.0552 \\
\bottomrule
\end{tabular}
\end{table}

\begin{figure}[!h]
    \centering
    \begin{subfigure}{\linewidth}
        \centering
        \vspace{6pt}
        
        \begin{tabular}{p{0.24\linewidth} p{0.35\linewidth} p{0.24\linewidth}}
            \centering \textbf{\small Noisy Input} & 
            \centering \textbf{\small Baseline Model} &
            \centering \textbf{\small Ground Truth} \par \vspace{1pt}
        \end{tabular}
        
        \vspace{-2pt}
        \includegraphics[width=\linewidth, trim=0 10 0 95, clip]{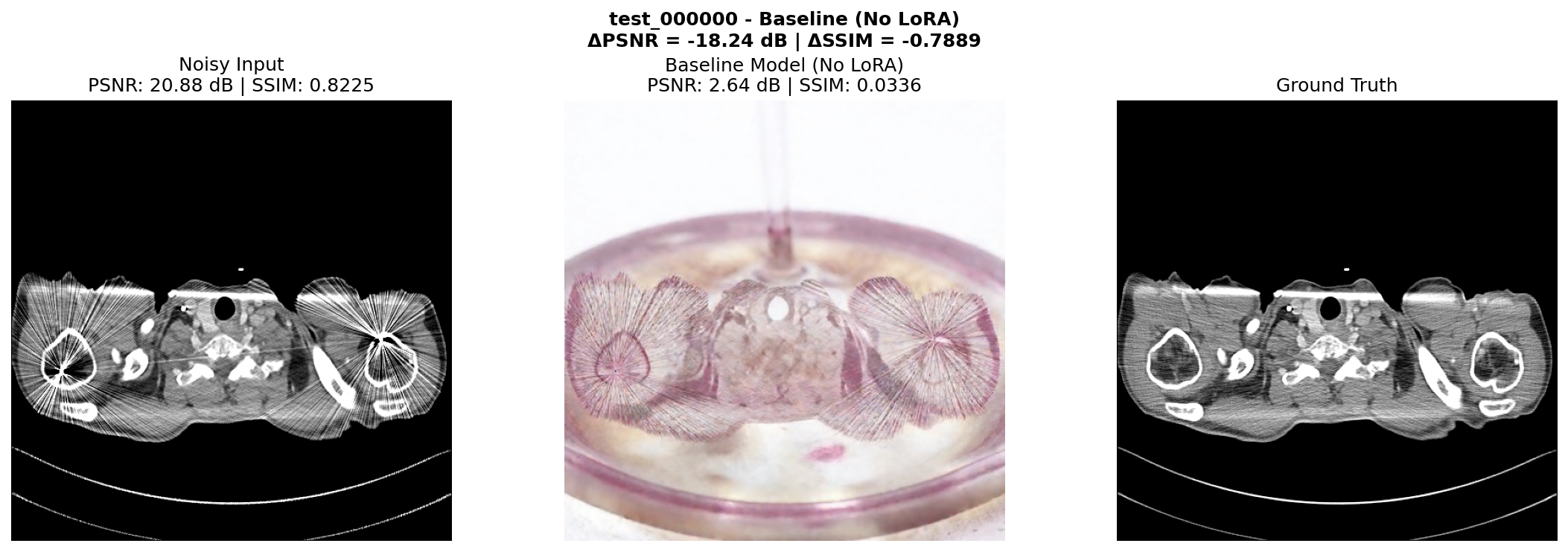} 
        \caption{Baseline model output showing a hallucinated petri dish.}
        \label{fig:baseline_pizza}
    \end{subfigure}

    \vspace{1.5em} 

    \begin{subfigure}{\linewidth}
        \centering
        \vspace{6pt}

        \begin{tabular}{p{0.24\linewidth} p{0.35\linewidth} p{0.24\linewidth}}
            \centering \textbf{\small Noisy Input} & 
            \centering \textbf{\small Baseline Model} &
            \centering \textbf{\small Ground Truth} \par \vspace{1pt}
        \end{tabular}

        \vspace{-2pt}
        \includegraphics[width=\linewidth, trim=0 10 0 95, clip]{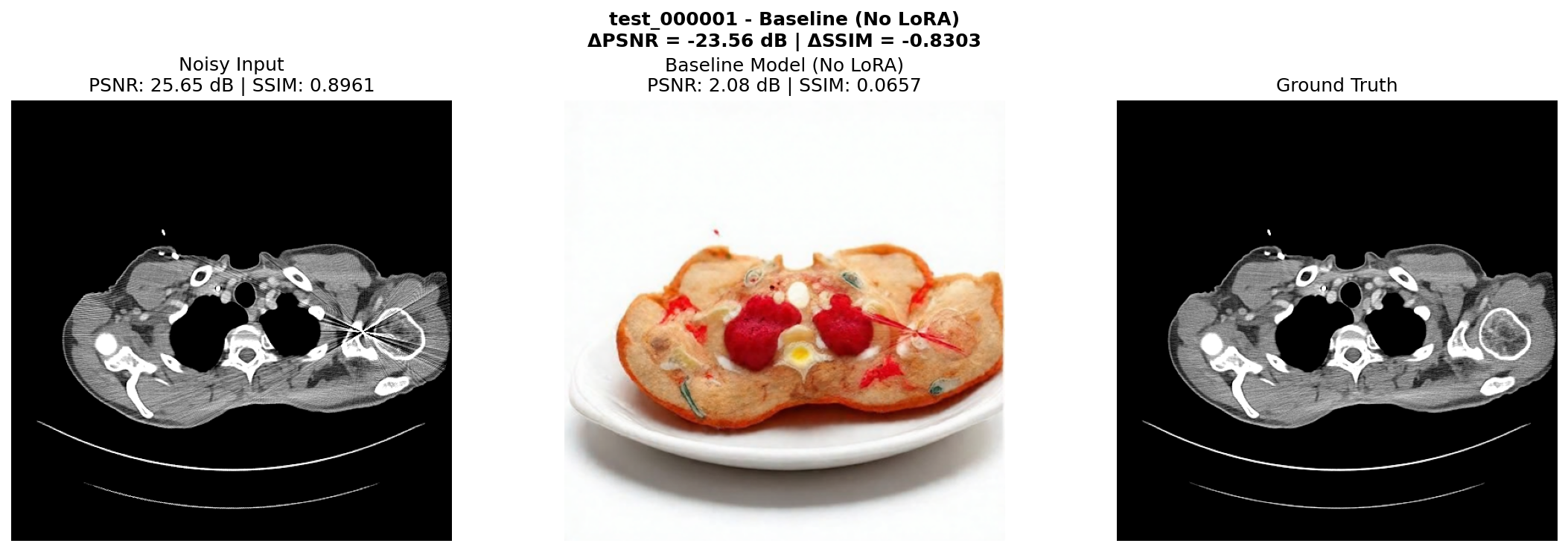} 
        \caption{Baseline model output showing a hallucinated waffle.}
        \label{fig:baseline_sofa}
    \end{subfigure}

    \caption{Visual demonstration of domain misalignment. Foundation models without LoRA adaptation hallucinate natural image features (petri dish, waffle) into CT reconstructions due to metal streak artifacts.}
    \label{fig:baseline_failure}
\end{figure}

\section{Results \& Discussion}
\label{sec:results}

\subsection{Ablation Study Setup}

We conduct two ablation studies to isolate the effect of LoRA rank $r$ and training set size $N$, fixing all other hyperparameters. Both experiments use the no-reference-conditioning setting to ensure a clean, controlled comparison.

\subsection{Need for Domain Adaptation}
Table~\ref{tab:main_results} reveals that, the zero-shot evaluation of the base Qwen-Image-Edit model without any LoRA fine-tuning yields severely degraded outputs, with a PSNR of just 4.21 dB, an FID of 193.66, and an LPIPS-Rad of 0.7609. As shown in Figure~\ref{fig:baseline_failure}, the model actively hallucinates natural image content into the reconstruction: metal streak artifacts, which radiate outward in spoke-like patterns visually similar to certain natural image textures, are interpreted as objects such as a petri dish or waffle (Figures~\ref{fig:baseline_pizza} and~\ref{fig:baseline_sofa}). These hallucinated structures could be mistaken for genuine anatomy, rendering the outputs non-diagnostic. Crucially, LoRA fine-tuning on as few as 16 paired examples eliminates this behavior entirely, underscoring that the bottleneck is domain alignment rather than representational capacity: the pretrained backbone already possesses the structural priors needed for high-quality restoration but requires targeted adaptation to apply them in the CT domain.

\begin{table}[h]
\centering
\caption{Reference conditioning strategy comparison at $N{=}128$, and $r{=}64$.
Methods: None, Grid = Grid-concat, Multi = Multi-image. \textbf{Bold}: best per metric.}
\label{tab:conditioning_comparison_transposed_singlecol}
\setlength{\tabcolsep}{2.5pt}
\renewcommand{\arraystretch}{1.10}
\begin{tabular}{lcccccc}
\toprule
Method
& \shortstack{PSNR}
& \shortstack{SSIM}
& \shortstack{FID}
& \shortstack{FID-Rad}
& \shortstack{LPIPS}
& \shortstack{LPIPS-Rad} \\
\midrule
None  & 31.61 & 0.8717 & 5.81 & 0.0038 & 0.1062 & 0.0437 \\
Grid  & 28.18 & 0.8256 & 7.31 & 0.0040 & 0.1083 & 0.0539 \\
Multi & \textbf{34.60} & \textbf{0.8829} & \textbf{4.91} & \textbf{0.0031} & \textbf{0.0926} & \textbf{0.0371} \\
\bottomrule
\end{tabular}
\end{table}

\subsection{VAE Ceiling}
We report the VAE round-trip ( encoding and decoding clean ground-truth images without any diffusion or LoRA) as the theoretical upper bound imposed by the lossy VAE bottleneck of the Qwen-Image-Edit backbone, standing at 38.57 dB PSNR, 0.9313 SSIM, and 0.0080 LPIPS-Rad. No method operating through this VAE can exceed these values regardless of adapter quality.

Our best ensemble configuration (ens-10) reaches 35.51 dB PSNR and 0.8966 SSIM, trailing OSCNet+ by 2.2 dB on PSNR. We attribute this gap primarily to OSCNet+'s use of sinogram-consistency supervision, which is unavailable to any purely image-domain method. Critically, on every perceptual and domain-aware metric our ensemble substantially outperforms all baselines, suggesting that OSCNet+'s PSNR advantage reflects regression-to-the-mean behaviour rather than superior anatomical fidelity. The remaining gap to the VAE ceiling is attributable to the inherent information loss of the VAE bottleneck rather than to limitations of the learned adapter.

\subsection{LoRA Rank Ablation}
Table~\ref{tab:rank_ablation} reports performance as a function of LoRA rank $r \in \{16, 32, 64, 128\}$ with $N{=}128$ training pairs. Performance improves substantially and monotonically from $r{=}16$ to $r{=}64$, with PSNR rising from 25.60dB to 31.61dB, SSIM from 0.7603 to 0.8717, and LPIPS-Rad from 0.1320 to 0.0437. At low ranks ($r \in \{16, 32\}$), the adapter lacks sufficient capacity to redirect the backbone's generative behaviour toward CT-domain restoration, producing outputs that retain visible streak artifacts. The sharp jump between $r{=}32$ and $r{=}64$ suggests the latter crosses a capacity threshold at which the adapter can simultaneously suppress metal-induced corruption and preserve surrounding anatomy.

Notably, increasing the rank further to $r{=}128$ yields a regression across all metrics (PSNR 30.49dB, FID 7.48, LPIPS-Rad 0.0552). We attribute this to the narrowness of the training distribution: with only $N{=}128$ pairs, the adapter is exposed to a highly limited range of anatomical configurations and artifact patterns, and a rank-128 adapter has sufficient capacity to overfit to this narrow distribution rather than learning a general restoration mapping that transfers to unseen test slices. Lower ranks act as an implicit regulariser, constraining the adapter to capture only the dominant modes of the restoration mapping that are reliably represented even in small training sets. We fix $r{=}64$ as our default for all subsequent experiments.

\subsection{Training Set Size Ablation}
Table~\ref{tab:size_ablation} reports performance as a function of training set size $N \in \{16, 32, 64, 128\}$ with rank $r{=}64$. Even at $N{=}16$ pairs the adapted model achieves LPIPS-Rad of 0.0649, already outperforming all three specialist baselines on this metric despite their training on orders of magnitude more data. This confirms that the foundation model's pretrained visual priors provide a powerful initialisation that partially substitutes for labelled supervision.

Performance improves consistently with $N$, with PSNR reaching 31.61dB and LPIPS-Rad 0.0437 at $N{=}128$, though marginal gains diminish at larger dataset sizes: the PSNR gain from $N{=}64$ to $N{=}128$ is only 0.20dB, compared to 1.27dB from $N{=}32$ to $N{=}64$. This saturation suggests the adapter approaches the limits of what $r{=}64$ can learn from the available supervision before $N{=}128$ is reached, and that further data scaling would yield diminishing returns without a corresponding increase in rank. We fix $N{=}128$ for all remaining experiments.

\subsection{Comparison with Baselines}

Table~\ref{tab:main_results} summarises the full comparison. We highlight the following observations per metric family.
PSNR and SSIM. OSCNet+ achieves the highest PSNR (37.74 dB) among all trained models, benefiting from its sinogram-consistency prior and training on a large synthetic dataset. Our ensemble variant achieves 35.51 dB, ranking second overall, while our single-sample configurations (31.61–34.60 dB) are comparable to ADN and RISE-MAR. The SSIM trend mirrors PSNR: our ensemble achieves the best SSIM (0.8966) of any trained model, including OSCNet+, indicating superior preservation of structural contrast and local texture.

\subsubsection{FID \& FID-Rad}

Our methods achieve dramatically lower FID and FID-Rad scores than all baselines. The best single-sample configuration (multi-refs, no ensembling) reaches FID 4.91 and FID-Rad 0.0031, compared to OSCNet+'s 31.60 and 0.0241 respectively corresponding near $6×\times$ improvement in distributional realism. This gap is striking given that OSCNet+ leads on PSNR, and underscores a fundamental tension in image restoration evaluation: pixel-fidelity metrics reward regression-to-the-mean behavior, which suppresses fine structural detail, whereas distributional metrics reward perceptual realism. Our foundation model approach avoids this regression by leveraging generative priors to produce outputs that lie within the manifold of real CT images.

\subsubsection{LPIPS and LPIPS-Rad.}
The perceptual distance results reinforce the FID findings.
Our multi-reference single-sample model achieves the best LPIPS (0.0926)
and LPIPS-Rad (0.0371) of any trained model, outperforming OSCNet+ by a
factor of approximately $1.7\times$ on LPIPS and $2.5\times$ on LPIPS-Rad.
The domain-aware LPIPS-Rad metric, which uses a ResNet50 pretrained on
165 radiological categories, is particularly informative for clinical
utility: it measures perceptual similarity in the representation space
used by downstream medical AI systems.
Our method's strong performance here suggests that the restored images are
not merely visually plausible but diagnostically consistent with real CT
anatomy.

\subsubsection{Reference Conditioning}
Table~\ref{tab:conditioning_comparison_transposed_singlecol} compares the two
reference conditioning strategies against the no-reference baseline.
Multi-image conditioning achieves the best results across all metrics,
most notably a PSNR of 34.60dB meaning a huge gain of 3.0dB over the
no-reference baseline alongside the lowest FID-Rad (0.0031) and
LPIPS-Rad (0.0371) of any single-sample configuration.

Grid-concatenation conditioning, by contrast, performs poorly, falling
below even the no-reference baseline on every metric despite having
access to the same five clean reference slices.
We attribute this to two compounding failure modes.
First, resizing each of the six images to $256{\times}256$ before tiling
reduces the effective resolution of both the artifact-corrupted input
and the reference slices by a factor of four relative to the native
$512{\times}512$ input, discarding precisely the fine-grained structural
detail that references are intended to supply.
Second, the spatial layout of the grid introduces ambiguity: the model
must infer from a text prompt alone which cell is the restoration target
and which are exemplars, relying on spatial understanding that was never
explicitly trained for this configuration.

In the multi-image conditioning strategy, by contrast, each image is encoded independently at full resolution and the prompt unambiguously labels the role of each token sequence, allowing the cross-image attention mechanism of the VLM backbone to draw structural cues directly and without resolution loss.

The failure of grid-concatenation therefore serves as an important negative result, demonstrating that simply providing reference images is insufficient and that the manner in which they are integrated into the model's context window is critical to whether anatomical cues can be exploited effectively.

\subsection{Self-ensembling}
To reduce stochastic variance at inference time, we apply self-ensembling over  independent samples, averaging the resulting pixel-space outputs (Section~\ref{sec:ensembling}). As shown in Table~\ref{tab:main_results}, this yields substantial pixel-fidelity gains: PSNR improves from 34.60dB to 35.51dB (+0.91dB) and SSIM from 0.8829 to 0.8966, making the ensemble the strongest image-domain trained model on both measures, within 3.1dB of the VAE ceiling. However, ensembling incurs a modest cost on distributional metrics (FID rises from 4.91 to 5.33 and LPIPS from 0.0926 to 0.0972) reflecting the known trade-off between variance reduction and perceptual sharpness. Which regime is preferable depends on the downstream task: ensembling suits quantitative workflows, while the single-sample configuration may be preferred when diagnostic sharpness is critical. Notably, even the ensembled outputs retain dramatically better perceptual scores than all specialist baselines.

\subsection{Qualitative Evaluation}
To stress-test each method under challenging conditions, Figure~\ref{fig:method_comparison} presents a heavily corrupted abdominal slice (noisy-input PSNR of 17.83dB) containing two high-attenuation metal implants connected by severe streak artifacts. ADN and RISE-MAR both attenuate streak intensity but leave clearly visible residual artifacts along the original trajectories, with little soft-tissue recovery in the artifact shadow. OSCNet+ achieves better streak suppression, yet faint directional artifacts and bright implant-region residuals persist.

Our method produces a visually clean reconstruction in which streak artifacts are almost entirely eliminated and the underlying abdominal anatomy is clearly delineated, with metal implant regions replaced by anatomically plausible tissue rather than bright residuals.
The absolute error maps (bottom row) corroborate these observations: baseline errors concentrate along streak trajectories, whereas ours are substantially lower and spatially uniform, confined primarily to sharp anatomical boundaries. Notably, all three baselines were trained on orders of magnitude more paired data than our 128-pair training set, and RISE-MAR \cite{risemar} in particular was trained on substantially milder corruption (~34,dB input PSNR), likely explaining its poor generalization to this severity level. Despite this data and distribution advantage, none fully resolves the artifact pattern suggesting that additional supervision alone does not compensate for the lack of strong generative priors under severe corruption.

\section{Conclusion}
We have presented a data-efficient framework for CT metal artifact reduction that adapts a general-purpose image editing foundation model via LoRA fine-tuning, requiring as few as 16 paired training examples to achieve competitive performance. At its core, our approach reframes artifact reduction as a multimodal reasoning problem: rather than learning a pixel-wise regression, the model is guided by a natural language instruction that communicates the intent of the restoration. This allows the model’s semantic understanding to inform where and how to intervene, effectively grounding its generative priors to prevent the anatomical hallucinations common in unadapted foundation models.

Multi-image reference conditioning extends this further, enabling the model to perform explicit in-context anatomical reasoning, comparing the artifact-corrupted input against clean exemplars to infer logically consistent tissue structures in affected regions. LoRA adaptation thus serves as a bridge, redirecting pretrained visual reasoning capabilities toward the clinical domain without discarding the structural priors that make that reasoning powerful. Our method substantially outperforms specialist baselines on perceptual and domain-aware metrics. These results suggest that the reasoning capabilities of vision-language models transfer effectively to clinical restoration, offering a scalable, reasoning-centric alternative to purpose-built networks for low-data deployment.

\paragraph{Limitations.}
Our evaluation is conducted entirely on the AAPM CT-MAR benchmark, which
relies on synthetically simulated metal implants rather than real
clinical acquisitions.
While this benchmark is widely adopted and reflects realistic artifact
patterns, the sim-to-real gap means that performance on genuine
clinical CT volumes remains to be verified.
Additionally, all experiments use a single foundation model
(Qwen-Image-Edit), so it is unclear how much of the observed benefit is
specific to this architecture versus general to instruction-conditioned
diffusion models.
Finally, the ensembling configuration requires ten forward passes at
inference time, which may be prohibitive in latency-sensitive clinical
settings without dedicated hardware.

\paragraph{Future work.}
The most pressing next step is validation on real clinical CT data, including a prospective reader study with radiologists to assess diagnostic utility beyond automated metrics. Natural extensions include applying the framework to other artifact types (motion blur, beam hardening, reconstruction noise), exploring alternative foundation model backbones to disentangle architectural contributions from pretrained priors, and extending to 3D volumetric inference to exploit inter-slice context. On the data side, active learning or uncertainty-guided sample selection could further reduce the annotation burden.
\clearpage
\bibliographystyle{ieeenat_fullname}
\bibliography{main}

\end{document}